%% file: bare_jrnl_new_sample4.tex
\documentclass[lettersize,journal]{IEEEtran}
\usepackage{amsmath,amsfonts}
\usepackage{algorithmic}
\usepackage{algorithm}
\usepackage{array}
\usepackage{subfig}
\usepackage{textcomp}
\usepackage{stfloats}
\usepackage{url}
\usepackage{verbatim}
\usepackage{graphicx}
\usepackage{cite}
\usepackage{multirow}
\usepackage{xcolor}
\usepackage{hyperref}
\usepackage{threeparttable}

\begin{document}
\title{Integrating Pruning with Quantization for Efficient Deep Neural Networks Compression}

\author{Sara Makenali, Babak Rokh, Ali Azarpeyvand*
\thanks{Sara Mekenali, Babak Rokh, and Ali Azarpeyvand are with the Department of Electrical and Computer Engineering, University of Zanjan, Zanjan, Iran (e-mail: sara.makenali@znu.ac.ir, babak.rokh@znu.ac.ir, and azarpeyvand@znu.ac.ir)}}

\maketitle

\begin{abstract}
Deep  Neural Networks (DNNs) have achieved significant advances in a wide range of applications. However, their deployment on resource-constrained devices remains a challenge due to the large number of layers and parameters, which result in considerable computational and memory demands. To address this issue, pruning and quantization are two widely used compression techniques, commonly applied individually in most studies to reduce model size and enhance processing speed. Nevertheless, combining these two techniques can yield even greater compression benefits. Effectively integrating pruning and quantization to harness their complementary advantages poses a challenging task, primarily due to their potential impact on model accuracy and the complexity of jointly optimizing both processes. In this paper, we propose two approaches that integrate similarity-based filter pruning with Adaptive Power-of-Two (APoT) quantization to achieve higher compression efficiency while preserving model accuracy. In the first approach, pruning and quantization are applied simultaneously during training. In the second approach, pruning is performed first to remove less important parameters, followed by quantization of the pruned model using low-bit representations. Experimental results demonstrate that our proposed approaches achieve effective model compression with minimal accuracy degradation, making them well-suited for deployment on devices with limited computational resources.
\end{abstract}


\section{Introduction
}
Deep Neural Networks (DNNs) have demonstrated significant capabilities in learning patterns and addressing a wide range of real-world challenges  \cite{wani2025advances}. However, they often require considerable computational resources for both training and inference. This can pose challenges in terms of time, memory, and energy consumption, especially on devices with limited hardware resources. To address these challenges, techniques such as approximate computing \cite{10816697, danopoulos2025transaxx} and model compression have been proposed to make these networks more efficient. Common compression techniques include quantization \cite{li2019additive, yvinec2023nupesnonuniformposttraining, bozorgasl2024clippeduniformquantizerscommunicationefficient, fu2025quantization}, pruning \cite{lee2018snip, singh2023efficientcnnspassivefilter, zu2023consecutive, khan2025pruning}, low-rank approximation \cite{li2023losparse, cherukuri2025low}, and Knowledge Distillation (KD) \cite{hemmatian2024uncertainty, lan2025counterclockwise}. By compressing DNNs, these networks become more efficient to deploy on resource-constrained devices, enhancing their practicality for real-world applications. The primary objective of compression techniques is to reduce the size of DNNs while preserving their predictive accuracy.

Two commonly used compression techniques are pruning and quantization. Pruning reduces the size of a neural network by removing connections or neurons that have little impact on the model accuracy. By eliminating these redundant components, the network becomes simpler, leading to faster inference times and reduced memory consumption. Quantization involves reducing the precision of the numerical components in a neural network, mapping them from higher-precision floating-point numbers to lower bit-width representations, such as integers \cite{10.1145/3623402}. This reduction in precision decreases the model size and computational requirements, making it more efficient for deployment on resource-constrained devices.

By combining pruning and quantization, it is possible to achieve even higher compression rates while maintaining acceptable prediction accuracy \cite{fan2021hfpq, balaskas2024hardware, naveen2025optimized}. Pruning reduces the overall size of the network by eliminating less important connections, whereas quantization minimizes the size of weights and activations. Together, these techniques effectively decrease the size of model and improve computational efficiency by simplifying the operations required during training and inference. A challenge in using a combination of pruning and quantization is effectively integrating these two compression approaches to maximize their potential while maintaining model accuracy. 

In this paper, we propose two effective approaches for integrating pruning and quantization to compress Deep Convolutional Neural Networks (DCNNs) during the inference phase while maintaining high accuracy. The first approach performs pruning and quantization simultaneously during each training epoch, allowing the model to update its parameters in alignment with both compression strategies throughout the training process. The second approach separates the training process into three distinct stages: (1) training the full-precision network to convergence, (2) applying incremental filter pruning, and (3) performing Quantization-Aware Training (QAT) on the pruned model. Accordingly, the first method is referred to as Simultaneous Pruning and Quantization (SPQ), while the second method is called Post-Pruning Quantization (PPQ). Both approaches aim to significantly reduce the memory and computational requirements of DCNNs, enabling efficient inference on resource-constrained edge devices. For pruning, we employ structured filter pruning based on the Geometric Median (GM) criterion \cite{he2019filter}, a similarity-based approach that identifies and removes less important filters. For quantization, we adopt the Additive Powers-of-Two (APoT) method \cite{li2019additive}, which accurately computes quantization levels using power-of-two (PoT) numbers to improve computational efficiency. The main contributions of the paper are summarized as:
\begin{itemize}
    \item We integrate GM-based pruning and APoT quantization for effectively compressing DCNNs. The GM criterion identifies less important filters to remove, by selecting those with the smallest Euclidean distance to GM, targeting filters with overlapping representations. This pruning strategy reduces model complexity while preserving accuracy. In parallel, APoT quantization approximates full-precision weight distributions using additive combinations of PoT values. This design enables efficient quantization and allows high-cost multiplications in multiply-accumulate (MAC) operations to be replaced with low-cost bit-shift operations, significantly improving computational efficiency.
    \item We propose and examine two distinct approaches for integrating pruning and quantization. The first method, SPQ, performs pruning and quantization simultaneously during each training epoch. This approach allows the model to update weights according to both the pruned structure and the quantized values, ensuring consistency between the compression process and model convergence, thereby maintaining high accuracy. The second method, PPQ, separates pruning and quantization into two stages to avoid compressing the model all at once, which could result in accuracy loss. Pruning is performed incrementally to preserve the representational capacity of the model during training and to mitigate potential degradation in accuracy. Once pruning is complete, QAT is applied to the pruned model to adapt the quantized weights to the new, compact architecture and minimize accuracy degradation.    
    \item The experimental results demonstrate that our approaches achieve a high compression rate with approximately a $\times$15 reduction in model size and a significant decrease in Bit-Operations (BOPs). Specifically, in comparison with State-of-the-Art (SOTA) methods, PPQ achieves the highest accuracy on ResNet-32 and ResNet-110, whereas SPQ attains the highest accuracy on VGG-16. These findings confirm the effectiveness of our approaches in achieving efficient compression while maintaining high accuracy.
\end{itemize}

The rest of the paper is organized as follows. In Section \ref{sec:related-works}, we review related works. Section \ref{sec:methods} thoroughly discusses the proposed approaches for combining pruning and quantization. The experimental results are presented in Section \ref{sec:experiments}. Finally, Section \ref{sec:conclusion} provides the conclusion and recommendations for future work.

\section{Related Works}\label{sec:related-works}
In this section, previous works related to pruning, quantization, and the combination of these two techniques are elaborated.

\subsection{Pruning}
Pruning is one of the most common techniques for reducing the number of parameters in DNNs. Pruning methods can be divided into structured and unstructured approaches. In unstructured pruning, individual weight connections are removed, resulting in an irregular and sparse connectivity pattern within the model. In contrast, structured pruning removes some structure of weight connections, such as channels or filters. While unstructured pruning can lead to models that are challenging to deploy on hardware due to their irregular structure, structured pruning produces regular models, making them suitable for implementation on hardware accelerators. As a result, specialized libraries or hardware are not required for deploying structured pruning on accelerators \cite{wang2020differentiable}.

References \cite{lee2018snip, han2015learning, shi2024sparse} suggest unstructured pruning. In reference \cite{han2015learning}, a method is introduced that removes weights smaller than a predefined threshold. The process involves first training the network to learn important connections, then pruning the weights below the threshold, and finally retraining the network to adapt the remaining weights to the new pruned structure. The SNIP method \cite{lee2018snip} introduces an innovative approach called connection sensitivity, which highlights crucial connections within a network before training. This method employs variance scaling initialization to detect and remove redundant connections, achieving the desired level of sparsity. SOGP \cite{shi2024sparse} presents a two-step strategy for pruning neural networks. In the first step, a sparsely ordered set model is generated during pretraining, which helps preserve the network structure. The second step involves element-wise sparse regularization to identify sparse weights and define effective pruning criteria. This method applies a threshold to prune less important weights.

Many recent works \cite{singh2023efficientcnnspassivefilter, he2019filter, luo2017thinet, he2018soft, ZHENG2024127124, lian2025daar} have proposed filter pruning. In ThiNet \cite{luo2017thinet}, information from the subsequent layer is used to determine candidate filters for pruning in the current layer. for each layer, the performance of the next layer is evaluated for various inputs, and filters with the least correlation to the output of the next layer are selected for pruning. This process involves optimization algorithms, including gradient descent and LASSO. However, this method is computationally intensive and time-consuming, particularly when training deep networks. SFP \cite{he2018soft} gradually removes unimportant filters using a threshold and trains the model with the remaining filters. This technique prunes filters based on their importance by temporarily ignoring their contribution to the network output at the end of each iteration. In the subsequent iteration, the selected filters are updated and used during the model training process. Reference \cite{ZHENG2024127124} proposes a filter pruning criterion that focuses on the direct and indirect effect of filters on the layers for identifying their importance.

Several works propose similarity-based criteria for identifying weights to prune \cite{zu2023consecutive, he2019filter, liu2023eacp, wang2023progressive, li2025sfp}. An issue with magnitude-based filter pruning is that filter norms are often concentrated within a narrow range, making it difficult to determine an appropriate pruning threshold \cite{he2019filter}. In such cases, the filters selected for pruning may significantly affect model accuracy. To tackle this, the FPGM method \cite{he2019filter} uses the GM criterion to capture the shared information among all filters within a layer, pruning those that lie closer to the GM due to their higher similarity. The idea is that the information from pruned filters can be represented by the remaining filters in the same layer, thus minimizing the impact of pruning on model accuracy.

PLFP \cite{wang2023progressive} prunes filters in a progressive manner, where Pruning is conducted layer by layer, focusing on the local geometric features of each filter. This approach emphasizes analyzing the local geometric properties of each filter and their relationships with neighboring filters. Instead of immediately removing the selected filters, their weights are progressively scaled down, gradually pruning them over time. This approach helps preserve the feature representation of model throughout the pruning process. Additionally, filters that may have been incorrectly identified for removal can be recovered in subsequent iterations of training.

The CLCS method \cite{zu2023consecutive} leverages the similarities among filters in successive layers of a network. By examining the filters in the current layer that correspond to those in the subsequent layer, similar filters are identified. Analyzing these similarities across filters in different layers enables to determination of comparable segments within the network, helping to make decisions on whether to remove or optimize them. The cosine similarity metric is employed to measure the similarity between filters. Once these similarities are computed, filters are grouped into clusters based on their shared characteristics, which informs decisions about their removal or adjustment.

\subsection{Quantization}
Quantization is the process of compressing a neural network by mapping continuous values to a smaller set of discrete finite values, aiming to reduce bit-width while minimizing error. Since quantization is independent of the network architecture, many previous works have utilized it to compress DNN models \cite{li2019additive, yvinec2023nupesnonuniformposttraining, bozorgasl2024clippeduniformquantizerscommunicationefficient, DBLP:journals/corr/MiyashitaLM16, zhou2017incremental, meiner2025prom}.
Based on the distances between quantization levels, quantization can be categorized into two types: 1) uniform quantization and 2) non-uniform quantization. Uniform quantization is simpler to implement, while non-uniform quantization often results in lower error as the quantization levels are computed more precisely \cite{10.1145/3623402}. References \cite{bozorgasl2024clippeduniformquantizerscommunicationefficient, meiner2025prom, liu2024qllmaccurateefficientlowbitwidth} propose uniform quantization, while references \cite{li2019additive, yvinec2023nupesnonuniformposttraining, DBLP:journals/corr/MiyashitaLM16, zhou2017incremental, lo2023block} focus on non-uniform quantization.

BSFP \cite{lo2023block} introduces an innovative quantization method that integrates block-based quantization with subword scaling. This approach divides the input data into blocks, enabling more precise quantization of small-scale values. Each block is then quantized using a subword scaling technique, where the resulting values are represented in floating-point format to enhance numerical precision.

PROM introduces a quantization approach tailored for depthwise-separable convolutional networks. It applies uniform ternary quantization to the computationally expensive 1×1 pointwise convolutions, while employing uniform 8-bit precision for other convolution types. Activations across all convolution layers are quantized to 8-bit precision. By eliminating multiplications in the pointwise layers and leveraging efficient 8-bit integer additions, PROM significantly improves computational efficiency. 

One type of non-uniform quantization is PoT quantization, where a full-precision value is mapped to the nearest PoT number. making it well-suited for implementation on FPGA platforms \cite{10.1145/3623402}. In references \cite{DBLP:journals/corr/MiyashitaLM16, zhou2017incremental}, quantization levels are limited to PoT values, reducing computational overhead by enabling multiplications to be implemented using bitwise shift operations on hardware platforms.

Reference \cite{li2019additive} demonstrates that the weights of DNNs follow a bell-shaped distribution and introduces Additive Powers-of-Two (APoT) quantization, an efficient non-uniform method for quantization of weights and activations. In this approach, quantization levels are limited to sums of PoT numbers, resulting in high computational efficiency and compatibility with the bell-shaped distribution of weights. As bit-width increases, APoT provides a well-distributed range of quantization levels, offering higher resolution near the peak of the bell curve and lower resolution for values farther from the peak.

\subsection{Combination of Compression Approaches}
Some works combine multiple compression approaches to achieve greater reductions in parameters and computational complexity \cite{fan2021hfpq, balaskas2024hardware, kim2021pqk, hasan2023compressed, 10946020}. The PQK method \cite{kim2021pqk} combines pruning, quantization, and KD. It consists of two phases: in the first phase, pruning and quantization are applied during training, and in the second phase, the compressed model is fine-tuned using KD from the original model.

The Hardware Compatible Pruning Quantization (HFPQ) method \cite{fan2021hfpq} introduces a post-pruning quantization strategy that integrates layer-wise channel pruning with PoT quantization. The process begins with channel pruning, followed by network retraining. After pruning, specific weights are quantized to PoT values based on the Euclidean distance between the original and quantized weights. To improve accuracy, the network is retrained with full-precision weights, and finally, all weights are quantized.

Reference \cite{ma2021non} introduces a framework for joint weight pruning and quantization. This work evaluates the effectiveness of both unstructured and structured pruning methods in conjunction with quantization, focusing on storage and computational efficiency. The study aims to determine whether reducing weights in DNNs through unstructured pruning can enhance performance across various platforms.

Reference \cite{hasan2023compressed} introduces an algorithm for training and pruning based on Recursive Least Squares (RLS) optimization. After an initial training period, the algorithm inverts the input autocorrelation matrices, integrates them with the weight matrices, and prunes insignificant input channels or nodes layer by layer. Each pruning step targets only a small portion of the channels or nodes, minimizing potential negative impacts. Training of the pruned network then continues, with subsequent pruning iteration delayed until the pruned network regains its performance.

Reference \cite{balaskas2024hardware} introduces an automated framework for compressing DNNs using a hardware-based approach. This framework combines pruning and quantization techniques, significantly enhancing pruning efficiency by integrating fine-grained and coarse-grained pruning methods. A composite reinforcement learning operator learns the optimal pruning and quantization configuration for each layer.

The combination of pruning and quantization offers a powerful strategy for enhancing the compression of DNNs. While these techniques individually focus on reducing model size and minimizing energy consumption, their integration can lead to significant improvements in both areas. However, the approach for efficiently combining these two techniques, which uses their capabilities in compression while maintaining accuracy, is challenging. In this paper, we propose two novel approaches for effectively combining pruning and quantization. 

In the first proposed approach, pruning and quantization are applied simultaneously during the training process, ensuring that weight updating is compatible with the compressed model. In contrast, the second approach separates pruning and quantization in the training phase to compress the model gradually and does not miss the capacity of the model at once to preserve accuracy. This method at first employs incremental pruning across multiple stages to maintain the accuracy of the pruned model. After the pruning process is complete, quantization is conducted on the pruned model.

In both proposed methods, structured filter pruning is applied to the convolutional layers. Given the weight-sharing nature of convolutional layers, similarity-based pruning can be more effective than magnitude-based approaches, as it targets and removes filters that produce overlapping feature representations. Therefore, We adopt the GM criterion, a similarity-based approach that identifies less important filters based on their Euclidean distance from the geometric center of all filters within the layer. Additionally, structured pruning is well-suited for deployment on hardware accelerators due to its regularity and compatibility with efficient memory access patterns. Filter pruning also offers considerable parameter reduction. Since the number of filters in one layer equals the number of input channels in the next, removing filters in a given layer also eliminates the corresponding input channels in the subsequent layer.  For quantization, we utilize the APoT approach to quantize both weights and activations. This method is well compatible with distribution of weights and activations, and enables efficient computation by replacing costly floating-point multiplications with low-cost bit-shift operations. The integration of GM-based filter pruning with APoT quantization results in a highly compressed model with reduced computations and storage requirements, while maintaining high accuracy, making it suitable for deployment on resource-constrained devices.

\section{Methodology}\label{sec:methods}
\begin{figure}[t]
\centering
\includegraphics[scale=0.65]{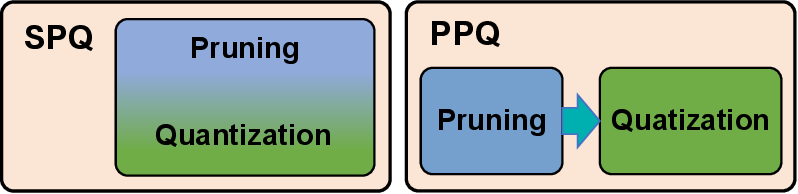}
\caption{Simultaneous pruning and quantization in SPQ, and sequential pruning and quantization in PPQ.}
\label{fig:method}
\end{figure}

In this section, the applied methods for pruning and quantization and the proposed approaches for integrating them are discussed. As illustrated in Fig.~\ref{fig:method}, the SPQ approach integrates pruning and quantization within a unified framework, applying both simultaneously during each training epoch. In contrast, the PPQ approach separates these processes: pruning is completed in the first phase, followed by quantization in the second phase. The remainder of this section first presents the pruning and quantization strategies used in the proposed methods, followed by a detailed description of each approach.

\subsection{Pruning}
In this work, we suggest a similarity-based filter pruning
approach, where filters that are highly similar to others are
identified as candidates for pruning. Convolutional layers utilize weight sharing, where each filter is convolved across the entire input feature map to extract features. However, some filters may learn highly similar representations. We hypothesize that pruning such similar filters, rather than relying on magnitude-based criteria, can improve pruning efficiency. By removing similar filters, the remaining filters can collectively approximate the representational capacity of the pruned filters. This approach preserves the overall feature extraction ability of the network and maintains its accuracy. To identify the filters most similar to others, the GM criterion is calculated and applied, as described in \cite{he2019filter}. GM estimates the center of data in Euclidean space. According to the Euclidean distance, the filters closest to GM in each layer are identified as pruning candidates. Equation~\eqref{eq:geometric} defines the computation of the GM for the filters in layer $i$.
\begin{equation}\label{eq:geometric}
\begin{aligned}
x^{GM} =\underset{x \in \mathbb{R}^{(C_{i}\times K_h\times K_w)}}{\arg \min } \sum_{j \in\{1,...,C_{i+1}\}}\left\|x-\mathcal{F}_{i, j}\right\|_2
\end{aligned}
\end{equation}
here, $x^{GM}\in \mathbb{R}^{(C_{i}\times K_h\times K_w)}$ denotes GM of filters. $C_i$, $K_h$, and $K_w$ indicate the number of input channels, kernel height, and kernel width, respectively. $C_{i+1}$ refers to the number of input channels in layer $i+1$, which corresponds to the number of filters (or output channels) in layer $i$. $\mathcal{F}_{i, j}$ represents the $j$-th filter in layer $i$. After computing the GM, the filters with the smallest distance to the GM are identified as candidates for pruning. The nearest filter to the GM is determined as Equation~\eqref{eq:distance}.
\begin{equation}\label{eq:distance}
\begin{aligned}
\mathcal{F}_{i, j^*} =\underset{\mathcal{F}_{i, j}}{\arg \min } \left\|\mathcal{F}_{i, j}-x^{GM}\right\|_2, \text { s.t. } j \in\{1,...,C_{i+1}\}
\end{aligned}
\end{equation}
Instead of computing Equations \eqref{eq:geometric} and \eqref{eq:distance}, we identify the filter with the minimum distance to other filters using the following equation, as proposed in \cite{he2019filter}: 
\begin{equation}\label{eq:GM}
\begin{aligned}
\mathcal{F}_{i, x^*}=\underset{x}{\arg \min } \sum_{j \in\{1,...,C_{i+1}\}}\left\|x-\mathcal{F}_{i, j}\right\|_2,\\
\text { s.t. } x \in\{\mathcal{F}_{i, 1},...,\mathcal{F}_{i, C_{i+1}}\}
\end{aligned}
\end{equation}

To perform pruning, a specific pruning ratio is defined, and the corresponding number of filters closest to the GM are selected for removal. By pruning these redundant filters, the remaining filters, being sufficiently diverse, retain the representational capacity of the network and can compensate for the pruned ones, thereby preserving the model accuracy.

\subsection{Quantization}
\begin{figure*}[t]
    \centering
    \subfloat[AlexNet on CIFAR-10.]{\includegraphics[scale=0.28]{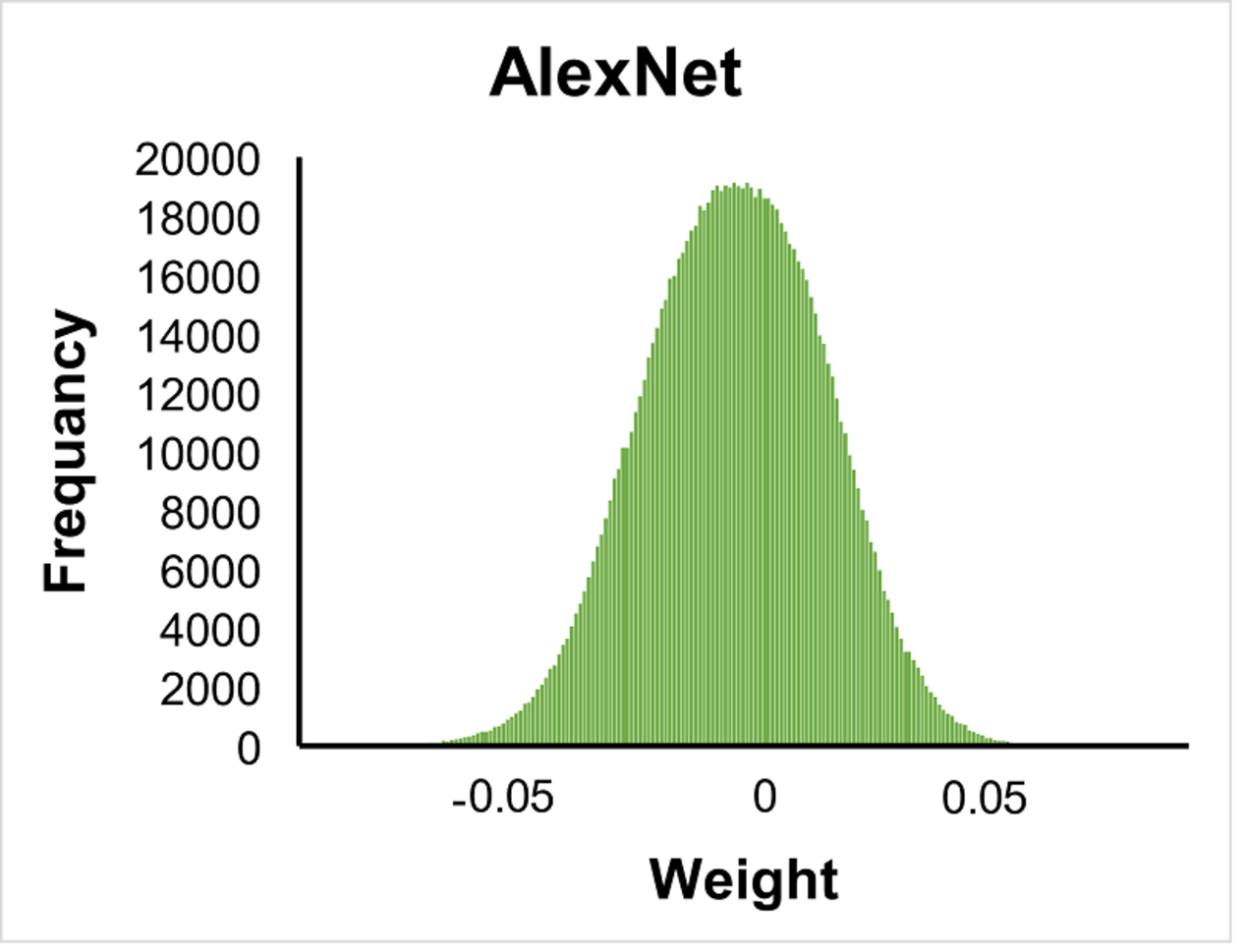}%
    \label{AlexNet-weights}}
    \hfil
     \subfloat[ResNet-20 on CIFAR-10.]{\includegraphics[scale=0.28]{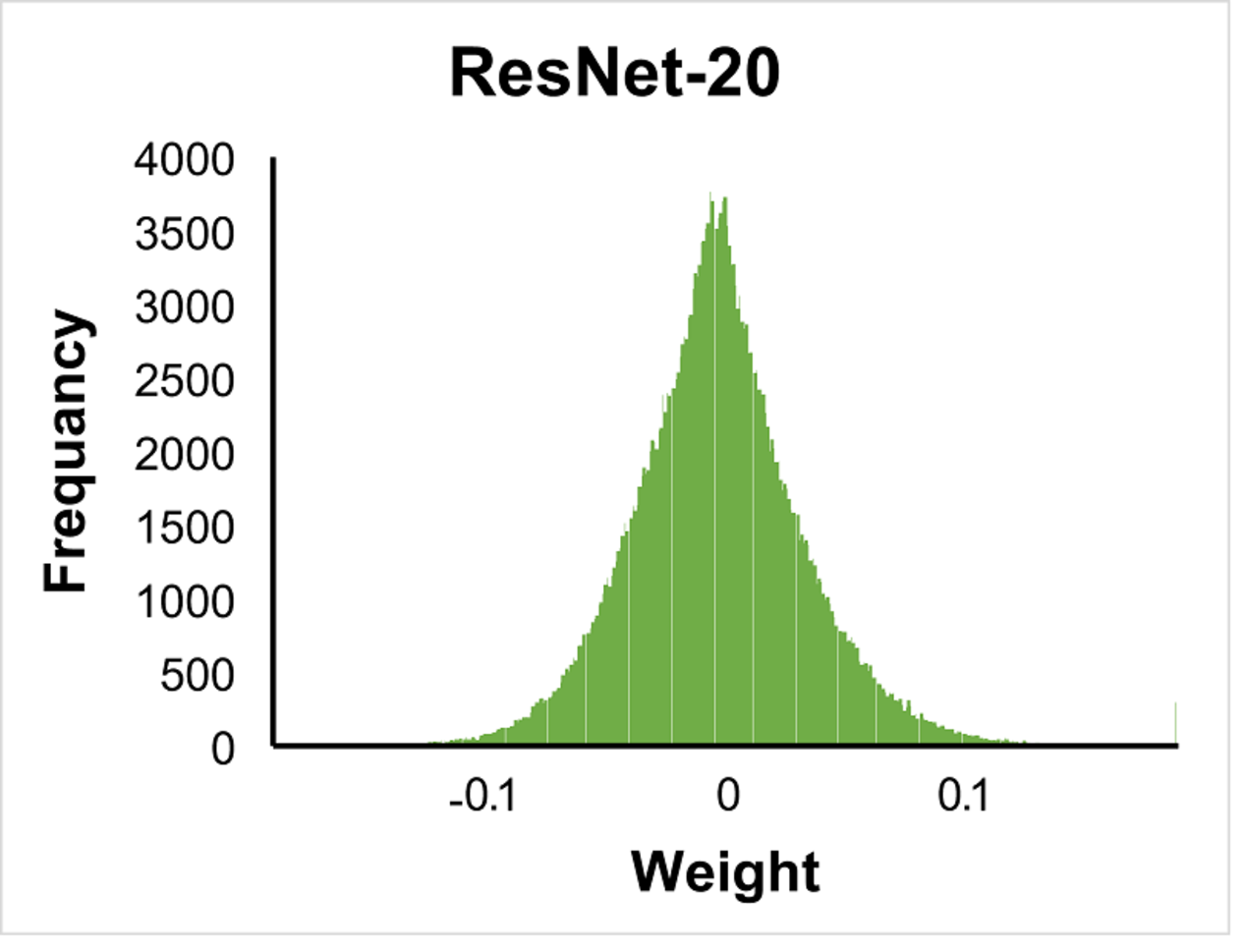}%
     \label{ResNet20-weights}}
    \hfil
     \subfloat[MobileNetV2 on ImageNet.]{\includegraphics[scale=0.28]{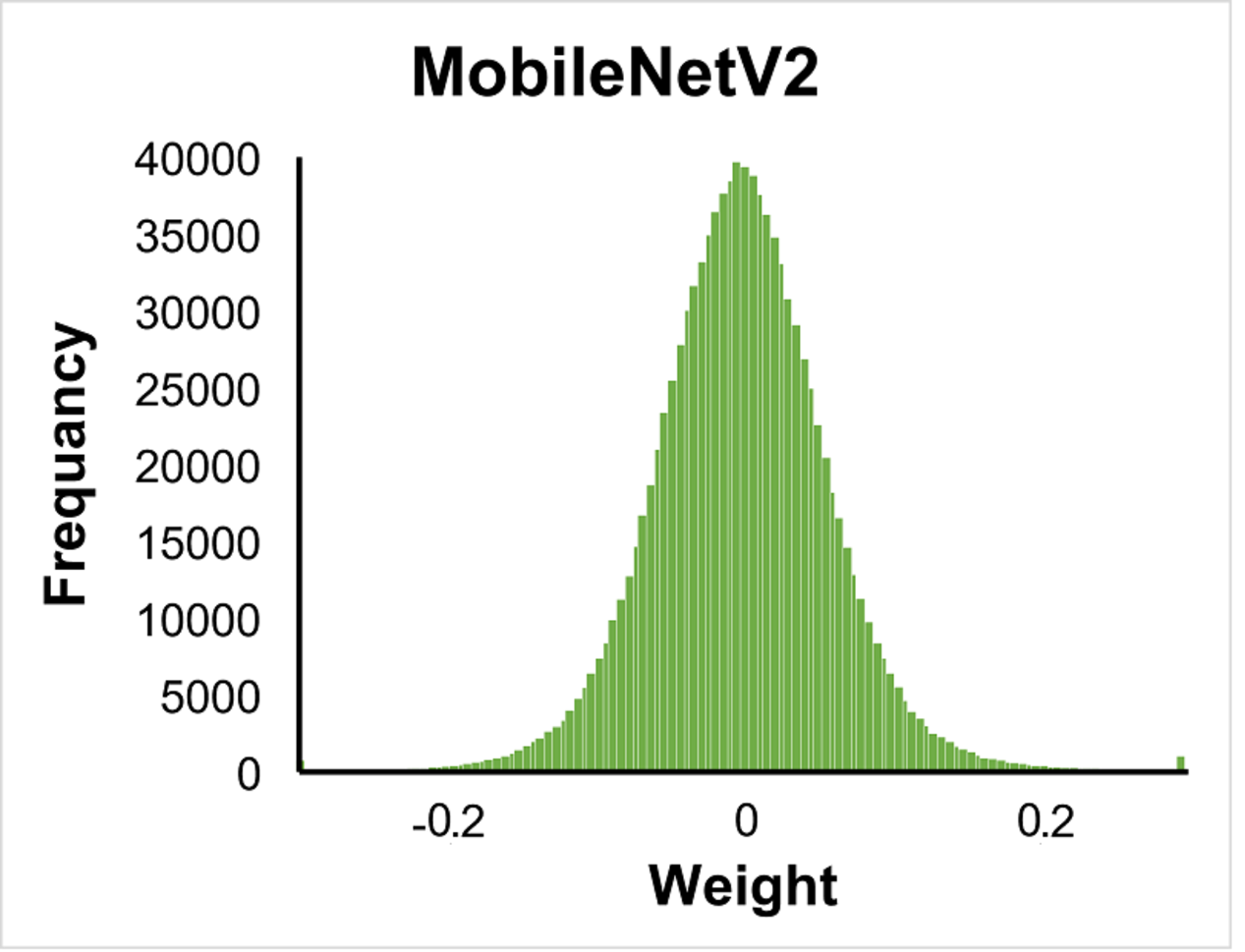}%
     \label{MobileNet-weights}}
    \caption{Weight distributions in trained models.}
    \label{weights-distribution}
\end{figure*}

Weights in a DNN typically follow a normal distribution with a mean near zero, resulting in most values being close to zero. Activations exhibit a similar pattern, though, due to the ReLU function, they are restricted to non-negative values \cite{DBLP:journals/corr/MiyashitaLM16}. For instance, Fig.~\ref{weights-distribution} shows the weights distribution of trained AlexNet and ResNet-20 on CIFAR-10, and trained MobileNetV2 on ImageNet. In PoT quantization, the quantization levels are powers of two, providing higher resolution near zero, where the peak of the weight distribution occurs, and lower resolution in the tails, where fewer weights are located.

A limitation of PoT quantization is that, as the bit-width increases, it primarily adds new quantization levels near zero, which leads to limited resolution improvement for weight values farther from zero. To address this issue, the APoT method \cite{li2019additive} has been proposed. Unlike PoT quantization, APoT enhances representational capacity by combining multiple PoT terms, producing a quantization level distribution that more closely matches the bell-shaped distribution of full-precision weights typically observed in DNNs. In this approach, while increasing the bit-width primarily adds most of the new quantization levels around the peak of the weight distribution, APoT also introduces new levels for values farther from zero. The quantization levels near the edges of the distribution are fewer compared to those near the peak, as the frequency of weights in the tails is lower. In APoT, each quantization level is calculated as the sum of $n$ PoT quantization levels according to Equation \eqref{eq:apot} \cite{li2019additive}.
\begin{equation}\label{eq:apot}
\begin{aligned}
& {Q}^{\alpha } (\alpha , {kn})= \gamma \sum_{i=0}^{n-1} P_i,\\
& \text { s.t. } \,
P_i \in    \left\{ 0, {\frac{1}{2^i}}, {\frac{1}{2^{i+n}}}, ..., {\frac{1}{2^{i+(2^k-2)n}}}\right\}
\end{aligned}
\end{equation}

where $P_i$ refers to a PoT quantization level. $\alpha$ specifies the clipping threshold, ensuring that the maximum quantization level does not exceed $\alpha$. $\gamma$ denotes the scale factor, while $k$ determines the base bit-width, which indicates the number of bits used in the base PoT quantization. $n$ represents the number of additive quantization levels in PoT quantization used to compute each quantization level in APoT, and it is defined as $n=b/k$, with $b$ being the bit-width of the APoT quantization. The number of quantization levels is given by $2^b=2^{nk}$.

For quantization, we adopt the APoT method due to its ability to achieve high representational capacity even at low precision. In both proposed compression approaches, weights and activations are quantized to 4-bit precision using APoT. This low-bit quantization significantly reduces memory requirements while maintaining accuracy. Moreover, floating-point multiplications in MAC operations can be efficiently replaced with bitwise shift operations \cite{li2019additive}. This substitution significantly enhances computational efficiency and reduces hardware complexity, making the quantized model more suitable for deployment on resource-constrained or edge devices.

\subsection{SPQ}
\begin{figure}[h]
\centering
\includegraphics[scale=0.36]{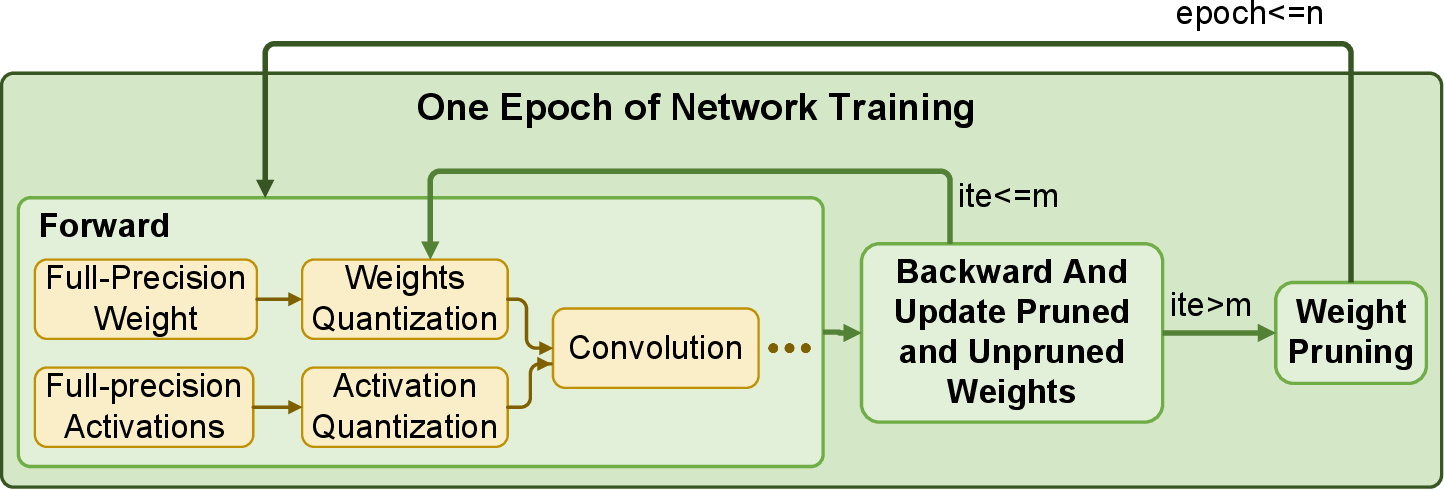}
\caption{An overview of the SPQ method.}
\label{fig:m1}
\end{figure}

In the SPQ method, during the training phase, weights and activations are quantized to 4-bit precision while simultaneously performing filter pruning in each epoch. As illustrated in Fig.~\ref{fig:m1}, the process begins with the quantization of weights and activations in each epoch. After quantization, both pruned and unpruned weights are updated, giving the pruned weights an opportunity to become effective in the subsequent epochs. Once $m$ iterations of quantization and updates are completed in each epoch, where $m=\frac{Training Samples}{BatchSize}$, pruning is applied at the end of the epoch. This process repeats for $n$ epochs.

\begin{figure*}[t]
\centering
\includegraphics[scale=0.31]{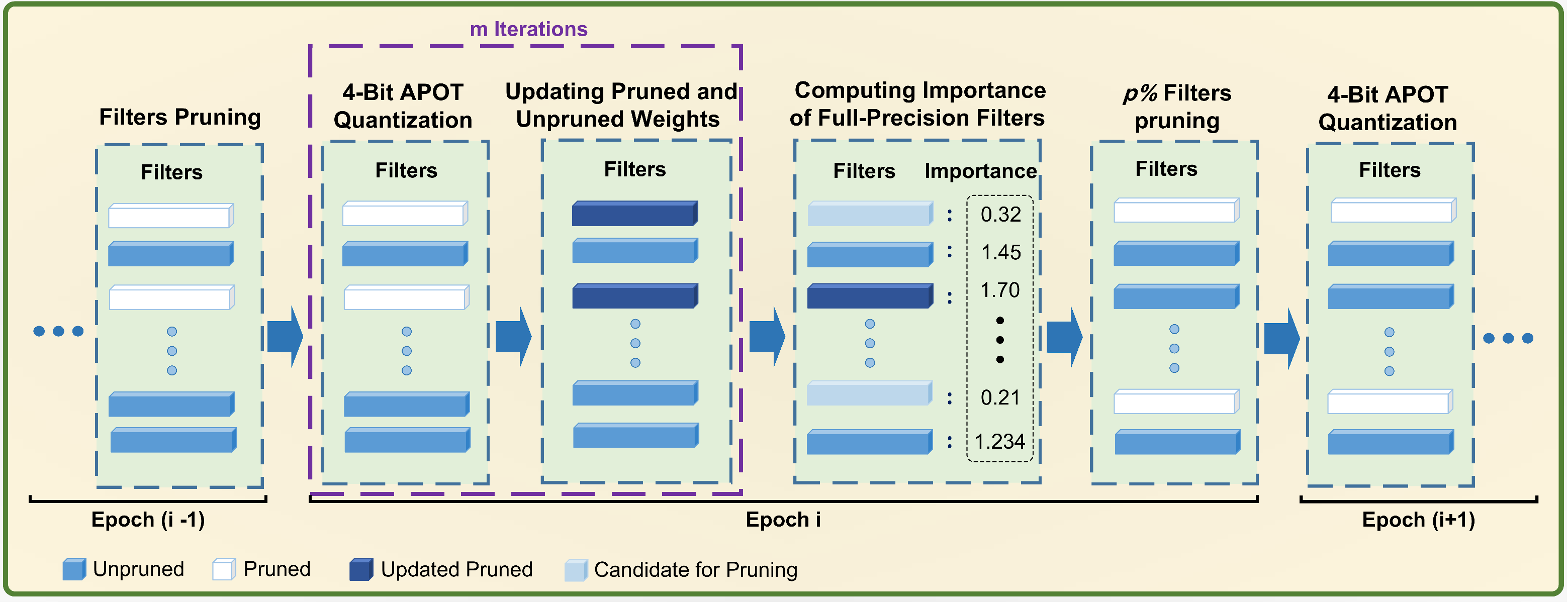}
\caption{The processing pipeline of weights in a training iteration for SPQ.}
\label{fig:m11}
\end{figure*}

Fig.~\ref{fig:m11} illustrates the process applied to the weights during training in the SPQ. In each epoch $i$, weights are quantized to 4-bit precision using the APoT method. After quantization, the weights are updated, with both unpruned and pruned weights from the previous epoch being updated. Updating the pruned weights allows them the chance to become effective in the subsequent epochs. Quantization and updating are performed over $m$ iterations. Afterward, pruning is carried out at the end of the epoch. Using the GM criterion, filters that contribute the least to model accuracy are identified as less important, and their weights are set to zero. It is important to notice that GM is computed based on full-precision weights. The pruned filters are then updated and included in the next epoch of training. During the forward pass, the contribution of the pruned filters to the network output is temporarily ignored. However, in the backward and update phases, the pruned filters are still allowed to be updated, giving them a chance to become effective. The pruned model from epoch $i$ is transferred to the subsequent epoch, and this iterative process continues until the quantized and pruned model converges.

In the SPQ method, quantization and pruning are integrated into each epoch of the training process, allowing the network to update its parameters in a manner consistent with both compression strategies. By performing pruning and quantization simultaneously, SPQ facilitates compression-aware learning throughout the entire training process, enabling the model to adapt more effectively to compressed representations.

Algorithm \ref{alg:alg1} provides the procedure for the SPQ method. In each epoch of training, quantization and weight updates are performed iteratively. At the end of the epoch, pruning is applied. Lines 4 and 5 apply APoT quantization to the activations and weights, respectively. The weights update is carried out on line 8 during each iteration. After completing all iterations, pruning is conducted in lines 11 to 17. Specifically, if the layer is a convolutional layer, GM is computed in line 13, and the $p_i\%$ of filters with the smallest distance to the GM are zero-masked in line 14, where $i$ denotes the index of the layer. Finally, the learning rate is adjusted in line 19.

\begin{algorithm}[h]
\caption{SPQ.}\label{alg:alg1}
\small
\begin{algorithmic}
\STATE \textbf{Input}: training data: $\mathcal{D}$, unpruned network: $\mathcal{W}$, number of layers: $L$, number of epochs: $n$, full-precision inputs: $X_{r}$, full-precision weights: $W_{r}$, pruning rate in layer $i$: $p_i$, initial learning rate: $\gamma$, batch size: $BS$. 
\STATE \textbf{Output}: compressed model: $\mathcal{W^*}$ where $\mathcal{W^*}<\mathcal{W}$. 
\STATE 1: For $epoch=1$ to $n$ 
\STATE 2: \hspace{0.2cm} For $iter=1$ to $\left\lceil\mid\mathcal{D}\mid/BS\right\rceil$
\STATE 3: \hspace{0.5cm} For $i=1$ to $L$
\STATE 4: \hspace{0.8cm} $X_{q}=APoTQuantization(X_{r})$
\STATE 5: \hspace{0.8cm} $W_{q}=APoTQuantization(W_{r})$
\STATE 6: \hspace{0.8cm} $Conv(X_{q}, W_{q})$
\STATE 7: \hspace{0.8cm} Update unpruned and pruned weights:
\STATE 8: \hspace{1cm} ${W_r}^{t+1}={W_r}^{t}-\gamma \frac{\partial{Loss}}{\partial{W_r}}$
\STATE 9: \hspace{0.6cm} End For
\STATE 10: \hspace{0.2cm} End For
\STATE 11: \hspace{0.2cm} For $i=1$ to $L$
\STATE 12: \hspace{0.5cm} If $i$ is a convolutional layer
\STATE 13: \hspace{0.8cm} Calculate GM for each filter
\STATE 14: \hspace{0.8cm} Zero-masking $p_{i}\%$ filters closer to the GM 
\STATE 16: \hspace{0.6cm} End If
\STATE 17: \hspace{0.2cm} End For
\STATE 18: \hspace{0.2cm} Update learning rate:
\STATE 19: \hspace{0.5cm} $\gamma^{t+1}=UpdateMethod(\gamma^{t},t+1)$
\STATE 20: End For
\end{algorithmic}
\label{alg1}
\end{algorithm}

\begin{figure*}[b]
\centering
\includegraphics[scale=0.35]{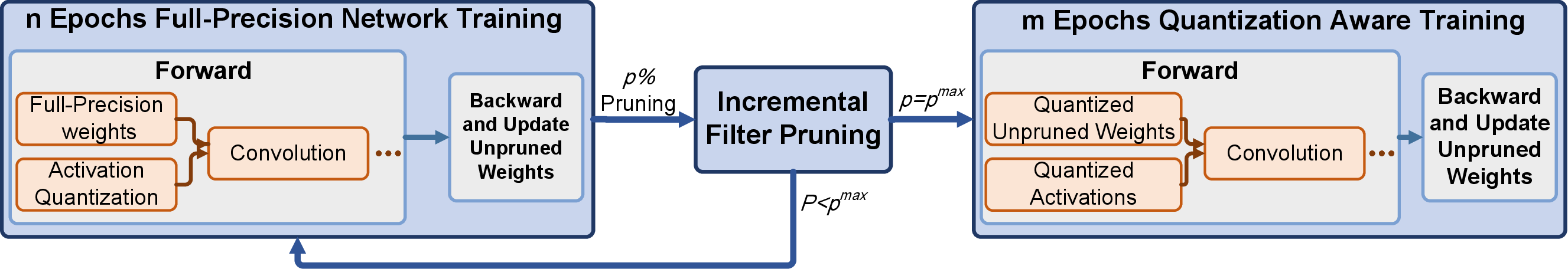}
\caption{An overview of the PPQ method.}
\label{fig:m2}
\end{figure*}

\subsection{PPQ}
In the PPQ method, we propose a combination of pruning and quantization to compress a DNN model, where pruning and quantization are conducted separately during training. Fig.~\ref{fig:m2} presents an overview of this method, in which incremental pruning is performed before quantization. Instead of pruning all filters at once, the pruning rate gradually increases. Initially, the full-precision network is trained for $n$ epochs, after which $p\%$ of the filters are pruned, where $p<p^{max}$, and $p^{max}$ is the final desired pruning rate. This process continues until the pruning rate reaches $p^{max}$. The gradient of the pruned filters is set to zero at each stage, meaning that, unlike the SPQ, the pruned filters are not restored and remain zeroed out. After pruning is complete, quantization of the activations and weights is performed over $m$ epochs using the QAT approach.

\begin{figure*}[t]
\centering
\includegraphics[scale=0.27]{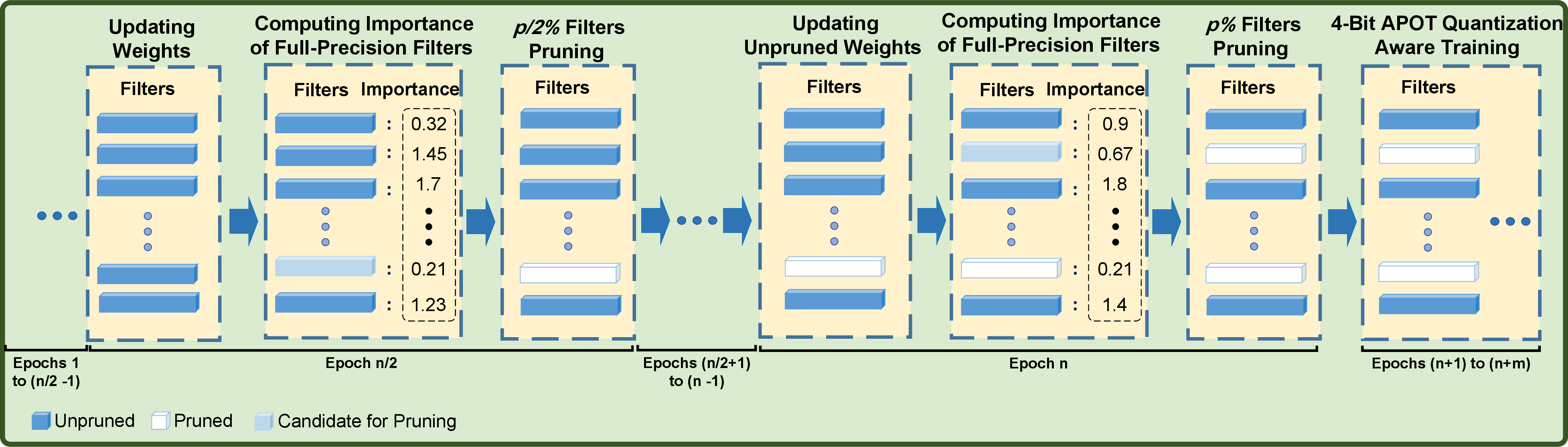}
\caption{Illustration of the weight pruning and quantization process throughout training in PPQ.}
\label{fig:m22}
\end{figure*}

Fig.~\ref{fig:m22} illustrates the process applied to the weights during training in the PPQ. Here, we assume the full-precision network is trained for $n$ epochs, with incremental pruning performed in two stages at a target pruning rate of $p\%$. After training for $n/2$ epochs, the pruning process begins by removing $p/2\%$ of the less important filters, as determined by the GM criterion. The network is then trained for another $n/2$ epochs, followed by pruning an additional $p/2\%$ of the filters, bringing the total pruning ratio to $p\%$. After completing the pruning process, quantization is performed using the APoT method over $m$ epochs of QAT. In this phase, the weights from the pruned model are utilized to maintain compatibility between quantization and the new architecture of the model. As a result, the updated quantized weights are well-matched to the pruned structure, enabling effective compression while preserving accuracy.

In PPQ, pruning is conducted incrementally over multiple steps. This gradual pruning strategy ensures that the network does not lose its representational capacity under high pruning rates. Instead, the model is given sufficient time during the subsequent training epochs to adapt and compensate for the removed filters, thereby mitigating accuracy degradation commonly associated with aggressive pruning. Furthermore, separating pruning and quantization enables more stable model compression, which contributes to enhanced accuracy.

Algorithm \ref{alg:alg2} outlines the process of PPQ. The algorithm consists of two distinct phases: the first part corresponds to pruning, while the second part performs quantization. Lines 1 to 20 are dedicated to the pruning phase. $n$ specifies the total number of epochs allocated to the pruning phase, and 
$s$ indicates the number of stages in which incremental pruning is applied. When the pruning module is called, GM is computed in line 12, and in line 13, the $p_{\text{stage}}$ fraction of filters with the smallest distance to the GM are zero-masked. The pruning ratio $p_{\text{stage}}$ increases in subsequent pruning stages. After pruning, quantization is carried out in lines 21 to 33. Initially, in line 22, the pruned model $\mathcal{W}^p$ is loaded. Then, from lines 23 to 33, Quantization-Aware Training (QAT) is performed based on the APoT method.

\begin{algorithm}[!htbp]
\caption{PPQ.}\label{alg:alg2}
\small
\begin{algorithmic}
\STATE \textbf{Input}: training data: $\mathcal{D}$, uncompressed network: $\mathcal{W}$, number of layers: $L$, number of epochs for pruning: $n$, number of epochs for quantization: $m$, full-precision inputs: $X_{r}$, full-precision weights: $W_{r}$, number of stages in pruning: $s$, pruning rates set: $\{p_i\}_{i=1}^{s}$, initial learning rate: $\gamma$. 
\STATE \textbf{Output}: compressed model: $\mathcal{W^*}$ where $\mathcal{W^*}<\mathcal{W}$.
\STATE 1: Pruning:
\STATE 2: \hspace{0.2cm} $stage=0$
\STATE 3: \hspace{0.2cm} For $epoch=0$ to $n-1$
\STATE 4: \hspace{0.4cm} Forward pass
\STATE 5: \hspace{0.4cm} Compute Loss and gradient in backward pass
\STATE 6: \hspace{0.4cm} Update unpruned weights:
\STATE 7: \hspace{0.5cm} ${W_r}^{t+1}={W_r}^{t}-\gamma \frac{\partial{Loss}}{\partial{W_r}}$
\STATE 8: \hspace{0.4cm} If ($epoch$ mod $(\lfloor n/s \rfloor)$ == 0)
\STATE 9: \hspace{0.6cm} $stage=stage+1$
\STATE 10: \hspace{0.5cm} For $i=1$ to $L$
\STATE 11: \hspace{0.8cm} If $i$ is a convolutional layer
\STATE 12: \hspace{1cm} Calculate distance to the GM for each filter 
\STATE 13: \hspace{1cm} Zero-mask $p_{stage}\% $ filters closer to the GM
\STATE 14: \hspace{0.8cm} End If
\STATE 15: \hspace{0.6cm} End For
\STATE 16: \hspace{0.4cm} End If
\STATE 17: \hspace{0.4cm} Update learning rate:
\STATE 18: \hspace{0.5cm} $\gamma^{t+1}=UpdateMethod(\gamma^{t},t+1)$
\STATE 19: \hspace{0.2cm} End For
\STATE 20: \hspace{0.2cm} $\mathcal{W}^p$ = pruned model
\STATE 21: Quantization:
\STATE 22: \hspace{0.2cm} Load $\mathcal{W}^p$
\STATE 23: \hspace{0.2cm} For $epoch=1$ to $m$ 
$\left\lceil\mid\mathcal{D}\mid/BS\right\rceil$
\STATE 24: \hspace{0.6cm} $X_{q}=APoTQuantization(X_{r})$
\STATE 25: \hspace{0.6cm} $W_{q}=APoTQuantization(W_{r})$
\STATE 26: \hspace{0.6cm} $Conv(X_{q}, W_{q})$
\STATE 27: \hspace{0.6cm} Update unpruned weights:
\STATE 28: \hspace{0.8cm} ${W_r}^{t+1}={W_r}^{t}-\gamma \frac{\partial{Loss}}{\partial{W_r}}$
\STATE 29: \hspace{0.6cm} Update learning rate:
\STATE 30: \hspace{0.8cm} $\gamma^{t+1}=UpdateMethod(\gamma^{t},t+1)$
\STATE 31: \hspace{0.2cm} End For
\end{algorithmic}
\label{alg2}
\end{algorithm}

\section{Experiments}\label{sec:experiments}
For evaluating the proposed methods, we use ResNet-20, ReNet-32, ResNet-56, ResNet-110 \cite{he2016deep}, and VGG-16 \cite{simonyan2014very} on the CIFAR-10 dataset \cite{krizhevsky2009learning}. The specifications of the uncompressed models for these networks, including the number of layers, test accuracy, and weight storage size in megabytes, are summarized in Table~\ref{tab1}. 
\begin{table}[!h]
	\centering
    \caption{Number of layers, test accuracy, and model size (in megabytes) of uncompressed networks on CIFAR-10.}   
	\begin{tabular}{@{}lccc@{}}
		\hline
		Model & Depth  & Accuracy& Model Size (MB)\\
		\hline  
		ResNet20 & 20 &  92.61 & 1.08\\ 
		ResNet32 & 32 &  93.08& 1.86\\
		ResNet56 & 56  & 94.30 & 3.41\\
		ResNet110 & 110 & 94.50 & 6.89\\
		VGG-16 & 16 &  94.05 & 59.91\\
		\hline     
	\end{tabular} 
 \label{tab1}
\end{table}

CIFAR-10 is one of the most widely used datasets in computer vision, which includes 60,000 color images distributed across 10 classes, with 50,000 images for training and 10,000 for testing.

For compression, the filter pruning module is applied to the convolutional layers of all networks with a pruning rate of 30\%. All layers are quantized to 4-bit precision, except for the first and last layers, which are quantized to 8-bit precision. During training, mini-batch gradient descent with a Momentum of 0.9 is employed. Cross-entropy is used as the loss function, combined with L2 regularization. These settings are consistently applied in both of the proposed methods.

\subsection{Training Setting}
Hyperparameters during training are adjusted as follows:
\begin{itemize}
    \item \textbf{SPQ:} In the SPQ, the initial learning rate is set to 0.1 and is decreased by a factor of 0.9 every three epochs if the accuracy does not improve. The weight decay for all models is set to 5e-4. 
    \item \textbf{PPQ:} In the PPQ, training includes two stages: pruning and quantization. During the pruning stage, the initial learning rate is set to 0.1 and follows the same decay schedule as in the SPQ. After completing the pruning stage, quantization is conducted. In the quantization, the trained weights from the pruning stage are set as initial weights. Since only small adjustments are needed during quantized model training to achieve optimal weights, we decrease exploration and increase exploitation by reducing the learning rate to 0.01. The weight decay remains the same as in the SPQ.
\end{itemize}
The batch size is set to 128 for all models across both proposed methods.

\subsection{Compression Metrics}
In addition to accuracy, we evaluate the effectiveness of the proposed approaches in achieving compression of DCNNs. We employ compression metrics, including model size and BOPs. The model size reflects the memory efficiency of the compressed model and is calculated as the total number of bits required to store the weights of the network. It is defined as
\begin{equation}\label{eq:size}
\begin{aligned}
ModelSize(x)=\sum_{i=1}^{L} \mid W_{i} \mid \times Bitwidth(w_i)
\end{aligned}
\end{equation}
where $L$ indicates the total number of layers in model $x$, $\mid W_i\mid$ determines the number of weights in layer $i$, and $Bitwidth(i)$ refers to the bit-width of the weights in layer $i$. 

The BOPs metric, which demonstrates the computational cost, is defined as \cite{wang2020differentiable}
\begin{equation}\label{eq:bop}
\begin{aligned}
BOPs=\sum_{i=1}^{L} MACs(i) \times Bitwidth(w_i) \times Bitwidth(a_{i-1})
\end{aligned}
\end{equation}

$MACs(i)$ represents the number of MAC operations in layer $i$, while $Bitwidth(a_{i-1})$ refers to the bit-width of the activations from layer $i-1$, which act as the inputs to layer $i$. These metrics provide insight into the storage and computational efficiency achieved by the compression methods.

\subsection{SPQ Versus PPQ}
In SPQ and PPQ, we propose two different approaches for integrating pruning and quantization. SPQ applies simultaneous pruning and quantization, whereas PPQ first trains a pruned model and then performs Quantization. In this section we compare the SPQ and PPQ methods.

\subsubsection{Convergence}
\begin{figure*}[t]
    \centering
    \subfloat[ResNet-20.]{\includegraphics[scale=0.57]{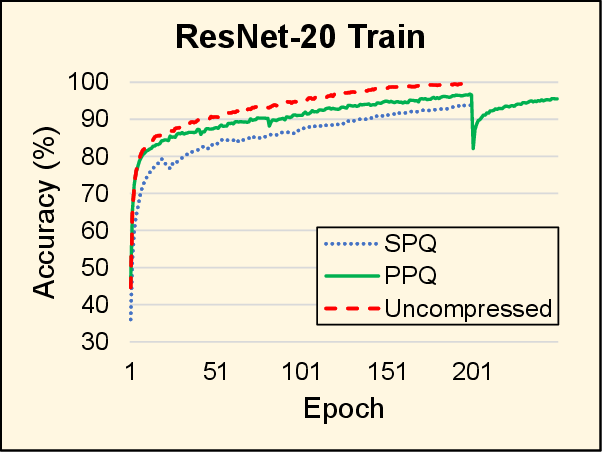}%
    \label{resnet-20-train}}
    \hfil
     \subfloat[ResNet-32.]{\includegraphics[scale=0.57]{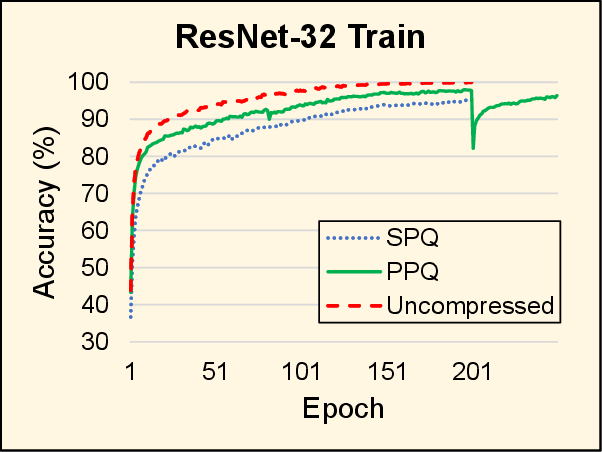}%
     \label{resnet-32-train}}
     \hfil
     \subfloat[ResNet-56.]{\includegraphics[scale=0.57]{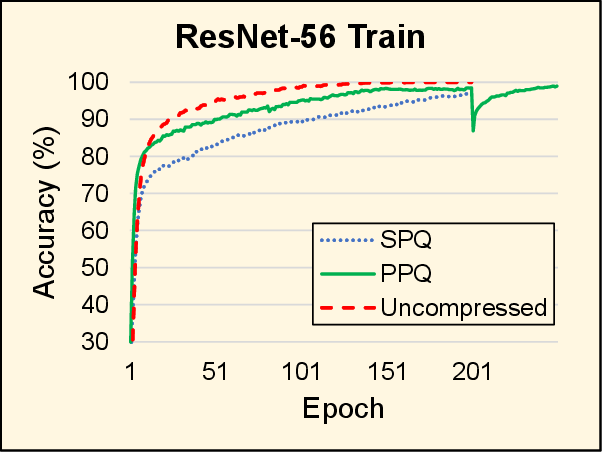}%
    \label{resnet-56-train}}
    \hfil
     \subfloat[ResNet-110.]{\includegraphics[scale=0.57]{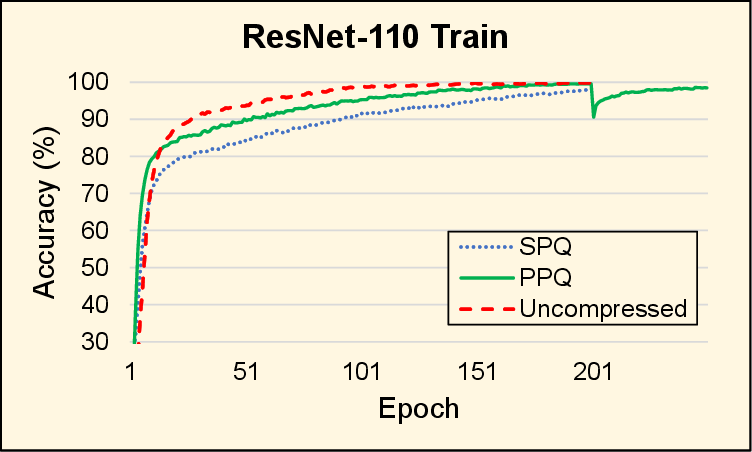}%
     \label{resnet-110-train}}
     \hfil
     \subfloat[VGG-16.]{\includegraphics[scale=0.57]{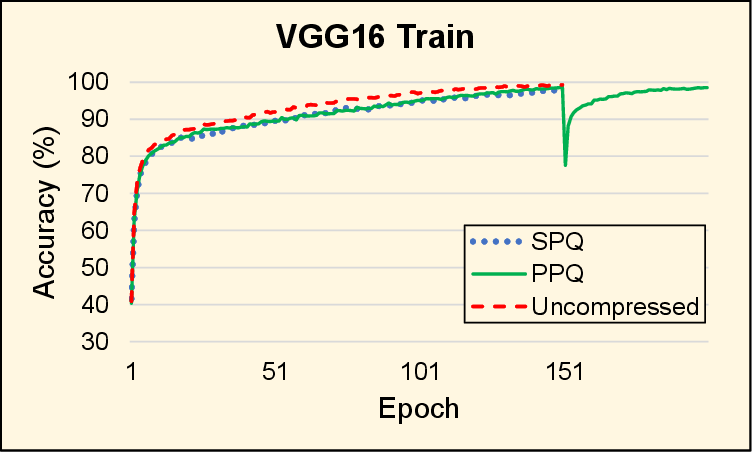}%
    \label{vgg-16-train}}
    \caption{Training accuracy convergence curves for SPQ, PPQ, and uncompressed models.}
    \label{fig:acc_curves}
\end{figure*}

Fig.~\ref{fig:acc_curves} represents the convergence curves during training for the uncompressed baseline model, SPQ, and PPQ on ResNet-20, ResNet-32, ResNet-56, ResNet-110, and VGG-16. The PPQ training process consists of two distinct phases. In the first phase, the pruned model is trained until convergence. In the second phase, this pruned model undergoes quantization and continues training using QAT. We refer to the first phase as pruning-PPQ.

For the ResNet architectures shown in Figures~\ref{resnet-20-train} to~\ref{resnet-110-train}, the pruning-PPQ model converges slightly faster than SPQ. SPQ, which applies pruning and quantization simultaneously, consistently converges around epoch 200 across all ResNet architectures. In contrast, the pruning-PPQ model converges earlier because it initially applies only pruning in its first phase. 

For VGG-16, as illustrated in Fig.~\ref{vgg-16-train}, the convergence pattern differs. All three models, the uncompressed model, pruning-PPQ, and SPQ, converge around epoch 150. 

After the first phase of training, PPQ proceeds with QAT. In this second phase, the pruned and quantized PPQ model converges after approximately 50 additional epochs. As a result, although the pruning-PPQ model tends to converge slightly earlier than SPQ during the initial phase, the overall convergence time for PPQ is longer due to the extra epochs required for training the quantized model.

\begin{figure}[h]
\centering
\includegraphics[scale=0.65]{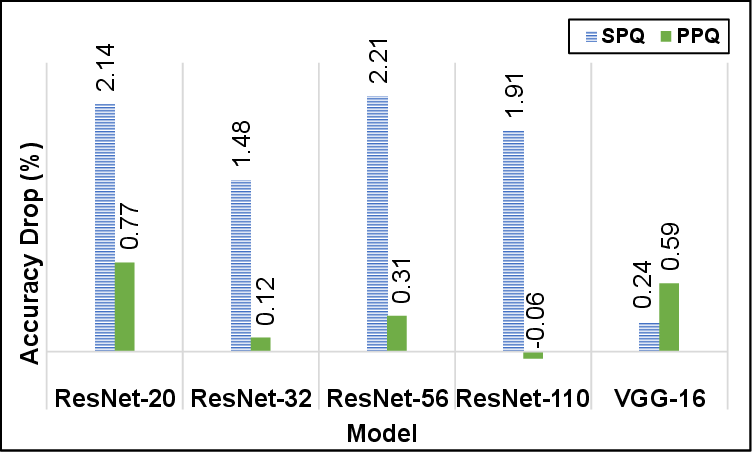}
\caption{Comparison of accuracy drop relative to the uncompressed model in the SPQ and PPQ methods.}
\label{acc}
\end{figure}

\subsubsection{Test Accuracy Comparison}
Fig.~\ref{acc} illustrates the accuracy drop of the models in SPQ and PPQ compared to the uncompressed model. The results show that PPQ yields a lower accuracy drop across all ResNet architectures. Notably, for ResNet-110, PPQ increases test accuracy by 0.06\%  compared to the baseline model. In contrast, for VGG-16, SPQ achieves superior accuracy, with an accuracy drop of only 0.24\%, compared to 0.59\% for PPQ.  

\subsection{Comparison with SOTA}
In this section, the proposed methods are compared with the uncompressed original model as the baseline and SOTA methods, including, APoT (4/4) \cite{li2019additive}, HFPQ (5/5) \cite{fan2021hfpq}, FPGM \cite{he2019filter},  SFP \cite{he2018soft},  BCGD (4/4) \cite{yin2019blended}, RAQ (2/2) \cite{li2021robustness}, SLB (4/4) \cite{yang2020searching}, BP-NAS (2.86/MP) \cite{yu2020search}, DCP \cite{chen2020dynamical}, ICP \cite{chang2023iterative}, HAP+IMPLANT \cite{yu2022hessian}, CTLP \cite{guenter2024concurrent}, SDN \cite{chen2018shallowing}, and IAP \cite{choudhary2022inference}. For all methods applying quantization, two numbers in parentheses are used: the first denotes the bit precision of the weights, and the second indicates the bit precision of the activations. In the BP-NAS method, MP refers to mixed precision. All tables are sorted in descending order of compressed accuracy, and the best result in each column is highlighted.

\subsubsection{Comparison on ResNets}
Table~\ref{tab_m1} presents a comparison of the ResNet models with depths of 20, 32, 56, and 110. As shown for ResNet-20, PPQ achieves an accuracy of 91.84\%, which is slightly lower than that of APoT, 92.30\%, and BP-NAS, 92.12\%. However, both of our methods achieve superior compression, with model size and BOPs reduction of $\times$15.78 and $\times$115.85, respectively. Although RAQ yields a higher compression rate than our methods, its accuracy of 89.80\% is lower than that of both PPQ, 91.84\%, and SPQ, 90.47\%.

For ResNet-32, PPQ achieves the highest accuracy among the compared SOTA methods, with only a 0.12\% drop compared to the uncompressed model. SPQ also maintains an acceptable accuracy of 91.59\%. Additionally, our approaches achieve the highest compression rates, with a $\times$15.84 reduction in model size and a $\times$118.18 reduction in BOPs.

PPQ achieves an accuracy of 93.99\%, only 0.01\% lower than the APoT method. Nevertheless, our proposed methods achieve the highest model size and BOPs reduction rates, reaching $\times$15.89 and $\times$119.88, respectively. SPQ also attains a satisfactory accuracy of 92.09\%. 

For ResNet-110, the PPQ approach achieves the highest accuracy of 94.56\%, which is 0.06\% higher than the baseline model. Moreover, our approach achieves the highest compression metrics, with a $\times$15.94 reduction in model size and a $\times$120.95 reduction in BOPs. SPQ attains an accuracy of 92.59\%, demonstrating acceptable performance given the achieved compression metrics.
\begin{table*}[!t]
	\centering
    \begin{threeparttable}
    \caption{Comparison with the uncompressed model as the baseline and SOTA methods on accuracy and compression metrics for ResNet architectures.}
	\begin{tabular}{@{}clccccc@{}}
		\hline
		Depth&Method&Baseline Acc.(\%)&Comp. Acc. (\%)&Acc. $\downarrow$ (\%)&Model Size $\downarrow$&
		BOPs $\downarrow$\\
		\hline
		\multirow{9}{*}{20}&APoT (4/4)\tnote{*}&91.60&\textbf{92.30}&\textbf{-0.7}&$\times$7.97&$\times$61.98\\
        &BP-NAS (2.86/MP)&92.61&92.12&0.49&$\times$10.74&$\times$95.61\\
        &\textbf{PPQ} (4/4)&92.61&91.84&0.77&$\times$15.78&$\times$115.85\\
        &BCGD (4/4)&92.41&91.65&0.85&$\times$8&$\times$64\\
        &SLB (4/4)&92.10&91.60&0.50&$\times$7.91&$\times$38.03\\        &SFP&92.20&90.83&1.37&$\times$1.98&$\times$1.89\\        &FPGM&92.20&90.62&1.58&$\times$2.65&$\times$2.58\\
        &\textbf{SPQ} (4/4)&92.61&90.47&2.14&$\times$15.78&$\times$115.85\\
        &RAQ (2/2)&91.80&89.80&2&$\times$\textbf{16}&$\times$\textbf{256}\\
        \hline
		\multirow{5}{*}{32}&\textbf{PPQ} (4/4)&93.08&\textbf{92.96}&0.12&$\times$\textbf{15.84}&$\times$\textbf{118.18}\\
        &DCP&92.36&92.25&\textbf{0.11}&$\times$1.94&$\times$2.04\\
        &SFP&92.63&92.08&0.55&$\times$1.99&$\times$1.90\\
        &FPGM&92.63&91.93&0.7&$\times$2.65&$\times$2.59\\
        &\textbf{SPQ} (4/4)&93.08&91.59&1.48&$\times$\textbf{15.84}&$\times$\textbf{118.18}\\
        \hline
        \multirow{10}{*}{56}&APoT (4/4)&93.20&\textbf{94.00}&\textbf{-0.80}&$\times$7.99&$\times$63.33\\
        &\textbf{PPQ} (4/4)&94.30&93.99&0.31&$\times$\textbf{15.89}&$\times$\textbf{119.88}\\
        &FPGM&93.59&93.49&0.10&$\times$2.66&$\times$2.60\\
        &SDN&93.03&93.29&-0.26&$\times$1.73&$\times$1.53\\
        &ICP&93.18&93.16&0.02&$\times$3.78&$\times$3.49\\
        &SFP&93.59&93.10&0.49&$\times$1.99&$\times$1.90\\
        &DCP&92.88&92.96&-0.08&$\times$2.04&$\times$2.04\\
        &HAP+IMPLANT&93.88&92.92&0.96&$\times$4.56&$\times$4.19\\
        &CTLP&94.26&92.53&1.73&$\times$2.07&$\times$3.29\\
        &\textbf{SPQ} (4/4)&94.30&92.09&2.21&$\times$\textbf{15.89}&$\times$\textbf{119.88}\\
        \hline
		\multirow{7}{*}{110}&\textbf{PPQ} (4/4)&94.50&\textbf{94.56}&-0.06&$\times$\textbf{15.94}&$\times$\textbf{120.95}\\
        &ICP&93.32&93.99&\textbf{-0.67}&$\times$4.81&$\times$3.49\\
        &FPGM&93.68&93.74&-0.14&$\times$2.66&$\times$2.60\\
        &DCP&93.38&93.54&-0.16&$\times$2.1&$\times$2.05\\
        &SFP&93.68&93.38&0.30&$\times$1.99&$\times$1.90\\
		&CTLP&94.70&93.02&1.68&$\times$3.39&$\times$4.83\\
        &\textbf{SPQ} (4/4)&94.50&92.59&1.91&$\times$\textbf{15.94}&$\times$\textbf{120.95}\\
		\hline     
	\end{tabular}
 \label{tab_m1}
 \begin{tablenotes}
   \item [*] The first number indicates the bit-precision of weights and the second number denotes the bit-precision of activations.
 \end{tablenotes}
 \end{threeparttable}
\end{table*}

\subsubsection{Comparison on VGGNet}
Table~\ref{tabvgg-m1} demonstrates the results for VGG-16. In terms of BOPs reduction, our proposed methods achieve a reduction of $\times$126.24, outperforming all other methods except HFPQ, which attains a higher reduction of $\times$309.70. Regarding model size reduction, our methods achieve a $\times$15.95 decrease, which is lower than those of HFPQ and ICP. However, both HFPQ and ICP exhibit lower accuracy compared to our proposed methods. Among all compared approaches, SPQ attains the highest accuracy of 93.81\%, while PPQ achieves an accuracy of 93.46\%, only 0.01\% lower than that of SDN.
\begin{table*}[b]
    \centering
    \caption{Comparison with the uncompressed model as the baseline and SOTA methods on accuracy and compression metrics for VGG-16.}
	\begin{tabular}{@{}lccccc@{}}
		\hline
        Method&Baseline Acc.(\%)&Comp. Acc. (\%)&Acc. $\downarrow$ (\%)&Model Size $\downarrow$&BOPs $\downarrow$\\
        \hline
        \textbf{SPQ} (4/4)&94.05&\textbf{93.81}&0.24&$\times$15.95&$\times$126.24\\
        SDN&93.50&93.47&\textbf{0.03}&$\times$8.26&$\times$1.64\\\textbf{PPQ} (4/4)&94.05&93.46&0.59&$\times$15.95&$\times$126.24\\
        ICP&93.60&93.29&0.31&$\times$27.28&$\times$7.25\\
        CTLP&94.05&92.74&1.31&$\times$11.83&$\times$2.72\\
        HFPQ (5/5)&93.58&92.47&0.65&$\times$\textbf{45.71}&$\times$\textbf{309.70}\\
        IAP&92.63&91.98&0.65&$\times$12.47&$\times$3.47\\
		\hline    
	\end{tabular} 
  \label{tabvgg-m1}
\end{table*}

Despite achieving high compression rates, the PPQ method consistently yields superior accuracy across all evaluated architectures, demonstrating the stability of this approach in combining pruning and quantization. In contrast, SPQ achieves competitive results on ResNet models and the highest accuracy on VGG-16, while requiring lower training cost than PPQ.

\subsubsection{Compression Efficiency}
\begin{figure*}[!t]
    \centering
    \subfloat[Model size for ResNet-20.]{\includegraphics[scale=0.68]{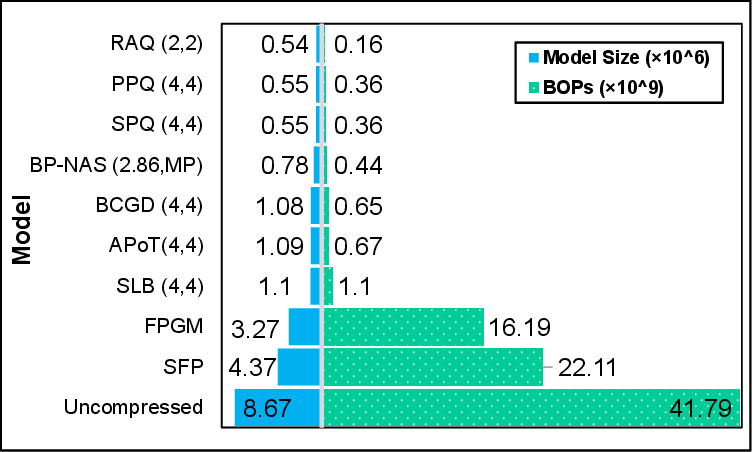}%
    \label{fig:resnet_20_size}}
    \hfil
     \subfloat[Model size for VGG-16.]{\includegraphics[scale=0.68]{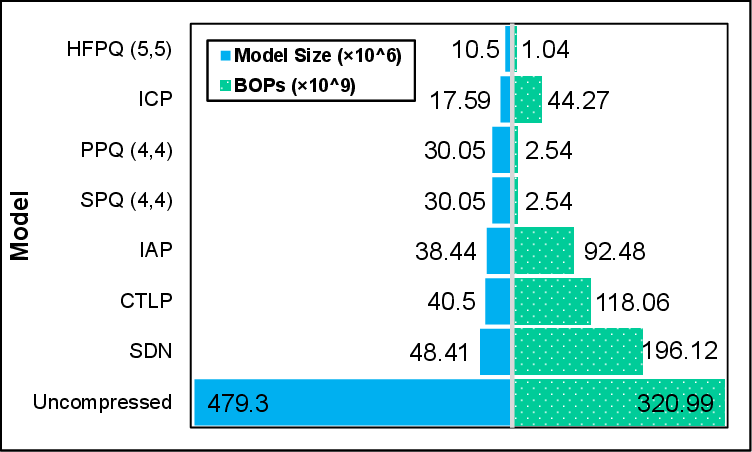}%
     \label{fig:VGG_16_size}}
     \hfil
    \caption{Comparison of model sizes and BOPs among the proposed methods, SOTA approaches, and the uncompressed model. The first number preceding each quantized model name indicates the bit precision of the weights, while the second indicates the bit precision of the activations.}
    \label{fig:model_size}
\end{figure*}

Considering the results, the proposed methods offer notable advantages in model compression by effectively combining pruning and quantization. For better visualization, Fig.~\ref{fig:model_size} compares the model sizes (in bits) and BOPs among the uncompressed model, our proposed methods, and various SOTA approaches on ResNet-20 and VGG-16. For ResNet-20, as illustrated in Fig.~\ref{fig:resnet_20_size}, both of our proposed approaches reduce the model size to 0.55$\times10^6$ bits and BOPs of 0.36$\times10^9$, representing a significant improvement compared to the uncompressed model, which has a size of 8.67$\times10^6$ bits and 41.79$\times10^9$ BOPs. The RAQ method further reduces the model size to 0.54$\times10^6$ bits, which is slightly smaller than that achieved by our methods. Moreover, RAQ attains BOPs of 0.16$\times10^9$ due to its use of 2-bit precision quantization for both weights and activations.

Fig.~\ref{fig:VGG_16_size} shows that our methods achieve a model size of approximately $30\times10^6$ bits and $2.5\times10^9$ BOPs on VGG-16, which are significantly smaller than the uncompressed model size of around $479\times10^6$ bits and $321\times10^9$ BOPs. Among the compared methods, HFPQ yields the smallest model, $10.5\times10^6$ bits, and the lowest BOPs, $1.04\times10^9$. Although the ICP method achieves a smaller model size of $17.59\times10^6$ bits compared to ours, it results in much higher BOPs, $44.27\times10^9$. This is because ICP relies only on high-rate pruning, which considerably reduces parameters and model size but has a limited impact on computational operations. In contrast, our approaches integrate pruning with quantization applied to both weights and activations, leading to a significant reduction in BOPs. It is also notable that all methods achieving higher compression than PPQ and SPQ in Fig.~\ref{fig:model_size} exhibit lower accuracy, as reported in Tables~\ref{tab_m1} and~\ref{tabvgg-m1}.

Since the proposed methods perform quantization of weights and activations using the APoT approach, floating-point multiplications in MAC units can be replaced with shift operations during hardware deployment of the compressed DNN. Consequently, in addition to reducing the number of parameters and computational operations, our approaches enable replacing high-cost floating-point multiplications with low-cost operations, thereby accelerating inference processing.

\section{Conclusion and Future Work}\label{sec:conclusion}
In this work, we proposed two novel methods for compressing DCNNs by integrating pruning and quantization. The SPQ method performs pruning and quantization simultaneously during training, and PPQ, sequentially applies pruning followed by quantization. Both methods employ the GM criterion as a similarity-based strategy for filter pruning, and employ the APoT quantization to efficiently represent weights and activations with low-bit precision. Experimental results demonstrated that integration of pruning and quantization in both approaches significantly reduces model size and BOPs compared to applying either technique individually. Among the two, PPQ consistently achieves superior accuracy across all tested architectures, outperforming several SOTA methods that apply only pruning or quantization. SPQ also performs competitively, particularly showing strong results on VGG-16. These outcomes confirm that our proposed approaches enable effective compression while maintaining high accuracy. For future work, pruning can be extended to the fully-connected layers , which typically contain a large proportion of network parameters. Moreover, exploring adaptive quantization strategies, such as layer-wise bit-width allocation and sensitivity-aware pruning rates, could further enhance the trade-off between compression and accuracy. 

\bibliographystyle{IEEEtran}

\input{bare_jrnl_new_sample4.bbl}



\end{document}